\documentclass[10pt,twocolumn,letterpaper]{article}

\usepackage{cvpr}              
\usepackage[accsupp]{axessibility}
\usepackage{multirow}
\usepackage{diagbox}
\usepackage{graphicx}
\definecolor{cvprblue}{rgb}{0.21,0.49,0.74}
\usepackage[pagebackref,breaklinks,colorlinks,allcolors=cvprblue]{hyperref}
\usepackage{cuted}

\usepackage{booktabs}
\usepackage[table]{xcolor}

\def\paperID{6385} 
\def\confName{CVPR}
\def\confYear{2026}

\title{BiPreManip: Learning Affordance-Based Bimanual Preparatory Manipulation \\ through Anticipatory Collaboration \\ }

\author{
Yan Shen \quad Feng Jiang \quad Zichen He \quad Xiaoqi Li \quad 
Yuchen Liu \quad Zhiyu Li \quad Ruihai Wu \quad Hao Dong \\
\vspace{2mm}
CFCS, School of Computer Science, Peking University \\
\vspace{3mm}
\href{https://sites.google.com/view/bipremanip}{
https://sites.google.com/view/bipremanip}
\vspace{-9mm}
}

\begin{document}
\maketitle

\begin{strip}
\vspace{-10mm}
    \centering
    \includegraphics[width=1\textwidth]{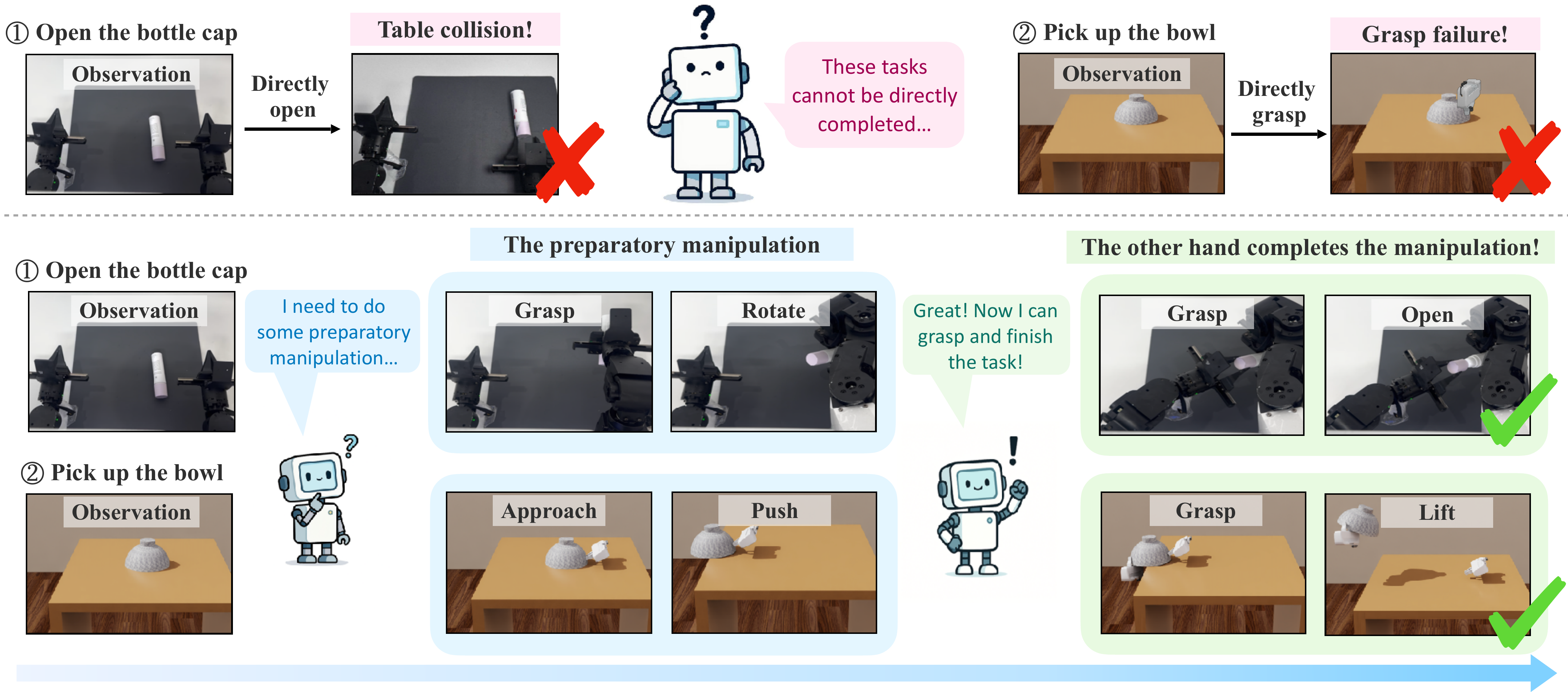}
    \vspace{-7mm}
    \captionof{figure}{
    Illustration of \textbf{Collaborative Preparatory Manipulation} tasks. (1) The top row shows objects (\emph{e.g.}, a capped bottle or an inverted bowl) that cannot be directly grasped or operated on by a single arm, highlighting the necessity of bimanual coordination and preparatory manipulation. (2) In the bottom rows, one arm first performs preparatory actions—such as lifting, reorienting, or repositioning an object—to enable the other arm’s subsequent goal-directed manipulation.
    }  
    \label{fig:teaser}
\end{strip}

\begin{abstract}
\vspace{-6mm}

Many everyday objects are difficult to directly grasp (\emph{e.g.}, a flat iPad) or manipulate functionally (\emph{e.g.}, opening the cap of a pen lying on a desk). Such tasks require sequential, asymmetric coordination between two arms, where one arm performs preparatory manipulation that enables the other’s goal-directed action—for instance, pushing the iPad to the table’s edge before picking it up, or lifting the pen body to allow the other hand to remove its cap.
In this work, we introduce \textbf{Collaborative Preparatory Manipulation}, a class of bimanual manipulation tasks that demand understanding object semantics and geometry, anticipating spatial relationships, and planning long-horizon coordinated actions between the two arms.
These tasks require the preparatory arm to act with awareness of the other arm’s goal, creating favorable object conditions for effective collaboration.
To tackle this challenge, we propose a visual affordance-based framework that first envisions the final goal-directed action and then guides one arm to perform a sequence of preparatory manipulations that facilitate the other arm’s subsequent operation. This affordance-centric representation enables anticipatory inter-arm reasoning and coordination, generalizing effectively across various objects spanning diverse categories. Extensive experiments in both simulation and the real world demonstrate that our approach substantially improves task success rates and generalization compared to competitive baselines. 

\end{abstract}    
\vspace{-6mm}
\section{Introduction}
\label{sec:intro}
\vspace{-1mm}

\definecolor{mydarkgreen}{HTML}{5a8a39}

\begin{figure}[t]
\centering
\includegraphics[width=\linewidth]{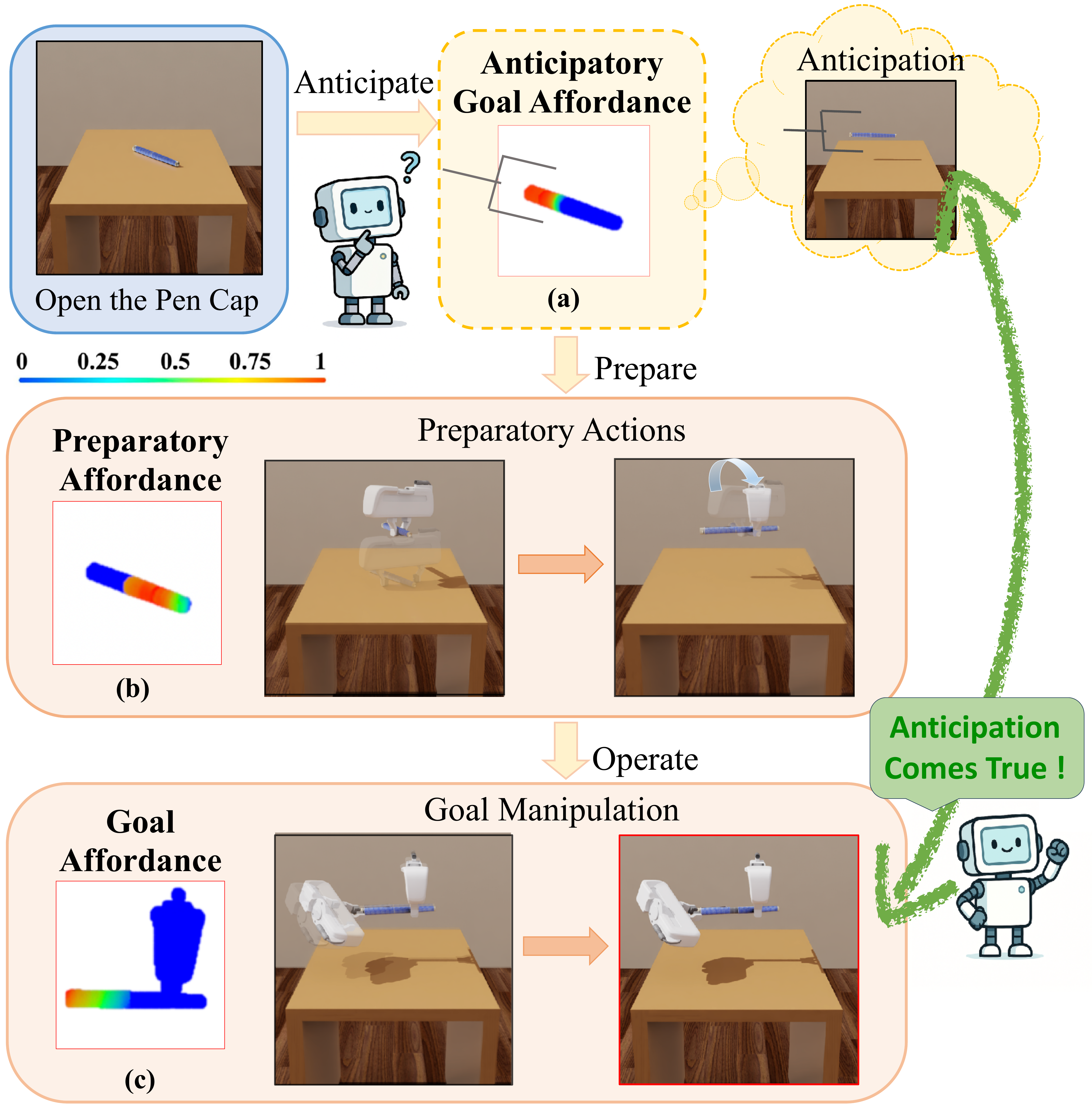}
\vspace{-7mm}
\caption{
\textbf{Overview of the BiPreManip framework} for bimanual preparatory manipulation tasks.
(a) The system predicts an anticipatory affordance map to infer the goal-directed interaction of the primary arm.
(b) Guided by this prediction, the assistant arm performs preparatory actions, establishing favorable conditions for manipulation.
(c) The primary arm then executes the goal-directed manipulation. 
This anticipatory reasoning enables effective and coordinated dual-arm manipulation.
} 
\vspace{-3mm}
\label{fig:intuition}
\end{figure}

Many everyday manipulation tasks involve objects that are difficult to grasp or manipulate directly, such as picking up a flat tablet or opening a capped pen lying on a desk. Unlike simple pick-and-place operations that can be accomplished by a single arm, these tasks often require one arm to first perform a series of preparatory actions, making the object accessible for the other arm’s goal-directed manipulation.
For example, as shown in Figure~\ref{fig:teaser}, one arm may lift a bottle's body to allow the other arm to remove its lid without colliding with the surface, or push an upside-down bowl toward the table edge so the other can grasp its rim. While such behaviors appear effortless for humans, they pose significant challenges for robots, requiring sequential coordination between two arms, perceptual understanding of object geometry and semantics, and long-horizon spatial reasoning over interdependent actions.

Recent advances in bimanual robotic manipulation have greatly enhanced dual-arm coordination through both algorithmic innovations~\cite{zhao2023learning, gkanatsios20253d, zhao2022dualafford} and benchmarking efforts~\cite{grotz2024peract2, mu2024robotwin, chen2025robotwin, zhang2024empowering, 10343126, mu2021maniskill}.
These studies have explored a wide range of collaborative behaviors, including symmetric coordination (\emph{e.g.}, jointly lifting or holding an object), sequential but independent subtasks (\emph{e.g.}, one arm opens a drawer while the other retrieves an item), and complementary yet directly executable roles (\emph{e.g.}, one arm pours from a bottle while the other holds a cup). However, most prior work assumes that both arms can directly interact with the object without first modifying its configuration or state.
In contrast, our work focuses on \textbf{Bimanual Preparatory Manipulation}, where one arm must first reconfigure the object to make subsequent manipulation by the other arm feasible. Such tasks demand asymmetric, anticipatory coordination and long-horizon interdependent planning, as the robot must reason about how each preparatory action establishes the physical conditions for the next. 
Without this capability, robots often struggle in scenarios where direct manipulation is initially infeasible.

To address this challenge, we introduce \textbf{BiPreManip}, a visual affordance-based framework for bimanual preparatory manipulation. BiPreManip first envisions the final goal-directed action through the affordance representation and then leverages the affordance to guide
preparatory actions in coordination with the other arm’s objective.
Specifically, BiPreManip first predicts how the primary arm—responsible for executing the goal-directed action—should ideally interact with the target object, generating an anticipatory affordance map that specifies where and how this interaction is expected to occur. Guided by this prediction, the assistant arm performs a sequence of preparatory actions, such as pre-grasping or reorienting the object, while avoiding interference with the anticipated interaction area, thereby enabling successful execution by the primary arm.
For instance, as illustrated in Figure~\ref{fig:intuition}, when a capped pen lies on its side, our framework anticipates that the target action is to open the cap; the assistant (right) arm first grasps and rotates the pen so that the cap faces the primary (left) arm, which then approaches and opens it.

To evaluate our approach, we establish a benchmark comprising diverse bimanual tasks that require preparatory actions. The benchmark includes a wide variety of objects from the ShapeNet~\cite{chang2015shapenet} and PartNet-Mobility~\cite{mo2019partnet,Xiang_2020_SAPIEN} datasets and 
encompasses several task types: (1) articulated objects must be lifted, rotated, or repositioned prior to functional manipulation of their movable parts, (2) flat or weakly graspable objects that must be pushed to the table edge for grasping, (3) a plate-lifting task adapted from PerAct2~\cite{grotz2024peract2}, where one arm presses one edge of a flat plate so the other can grasp the lifted side.
In addition, we conduct real-world robot–human handover experiments to validate BiPreManip’s applicability in robot-human collaborative settings, where the preparatory arm assists a human partner by facilitating pre-grasp configurations. 
Extensive experiments in both simulated and real-world environments demonstrate that BiPreManip substantially improves task success rates and generalization across diverse objects compared with existing baselines.

In summary, our main contributions are as follows:
\begin{itemize}
\item We introduce the problem of bimanual preparatory manipulation, where one arm performs preparatory actions to enable the other’s manipulation, emphasizing asymmetric coordination and long-horizon reasoning.
\item We propose BiPreManip, a visual affordance-based framework that predicts anticipatory affordance maps for the primary arm’s intended interaction and guides the assistant arm’s complementary preparatory actions.
\item We develop a benchmark of preparatory manipulation tasks and show through extensive simulation and real-world experiments that BiPreManip achieves superior performance and generalization across diverse objects.
\end{itemize}

\section{Related Work}
\label{sec:related}

\subsection{Bimanual Manipulation}
\vspace{-1mm}
Bimanual manipulation~\cite{zhao2023learning, chen2023diffusionpolicy, grannen2023stabilize, grotz2024peract2, liu2025factr,yuan2024anybimanual,aldaco2024aloha,fu2024mobile,ren2024enabling, zhou2025you, liang2026a3d, shen2025biassemble, li20253ds, zhao2022dualafford} enables coordinated dual-arm actions that demand broad spatial reach and stable inter-arm control.
Recent studies have advanced this area through improved coordination and representation learning. ACT~\cite{zhao2023learning} employs a transformer-based architecture that processes image inputs for bimanual action predictions via temporal chunks.
VoxAct-B~\cite{zhang2024voxactb} represents the workspace with voxel grids and decomposes the arms into acting and stabilizing roles, improving sample efficiency and generalization in bimanual tasks. InterACT~\cite{ren2024interact} introduces hierarchical attention to explicitly model inter-arm dependencies and synchronization.
At a larger scale, RDT-1B~\cite{liu2024rdt} introduces a diffusion foundation model that integrates vision and language for bimanual control.
More recently, 3D FlowMatch Actor~\cite{gkanatsios20253d} proposes a unified 3D policy for both single- and dual-arm manipulation using flow matching, achieving real-time inference while maintaining strong spatial grounding.  

In this paper, we introduce a class of sequential asymmetric collaborative tasks that require anticipatory inter-arm reasoning, termed the collaborative preparatory manipulation task. In this setting, one arm should first perform preparatory manipulation to reconfigure the object, creating the necessary affordance for the other arm’s subsequent goal-directed action, which requires spatial foresight and the avoidance of interference.
To address these tasks, we propose an affordance-based framework for learning sequential asymmetric bimanual collaboration that generalizes across various objects spanning diverse categories, achieving robust and 
coordinated dual-arm performance.

\subsection{Visual Representation for Manipulation}
\vspace{-1mm}
Visual representation plays a crucial role in bridging perception and action for robotic manipulation~\cite{nasiriany2025rt,bahl2023affordances,zhang2024affordance,zeng2021transporter,shridhar2023perceiver,shridhar2022cliport,do2018affordancenet,ding2024preafford,li2024manipllm}. It provides structured abstractions linking raw visual observations to manipulation control, enabling generalizable reasoning across diverse objects and tasks.
Early work focused on object-pose estimation~\cite{deng2020self,do2018deep} and part-segmentation pipelines~\cite{5152393,geng2023gapartnet}. 
More recent methods learn implicit object representations~\cite{wang2023d,qiu2025learning}, keypoints-based encodings~\cite{sundaresan2023kite,ai2025review}, or affordance-based representations~\cite{mo2021where2act,wu2021vat,wu2025garmentpile, shen2025biassemble} that capture functional and spatial properties critical for planning interactions with objects. Among these, affordance-based representations encode both the geometric and semantic structure of objects, explicitly serving as a bridge between perception and action. 
Building on this idea, our work extends such affordance-based visual representations to support anticipatory reasoning in sequential asymmetric bimanual preparatory manipulation tasks.

\section{Method}
\label{sec:method}

\begin{figure*}[t]
\centering
\includegraphics[width=\linewidth]{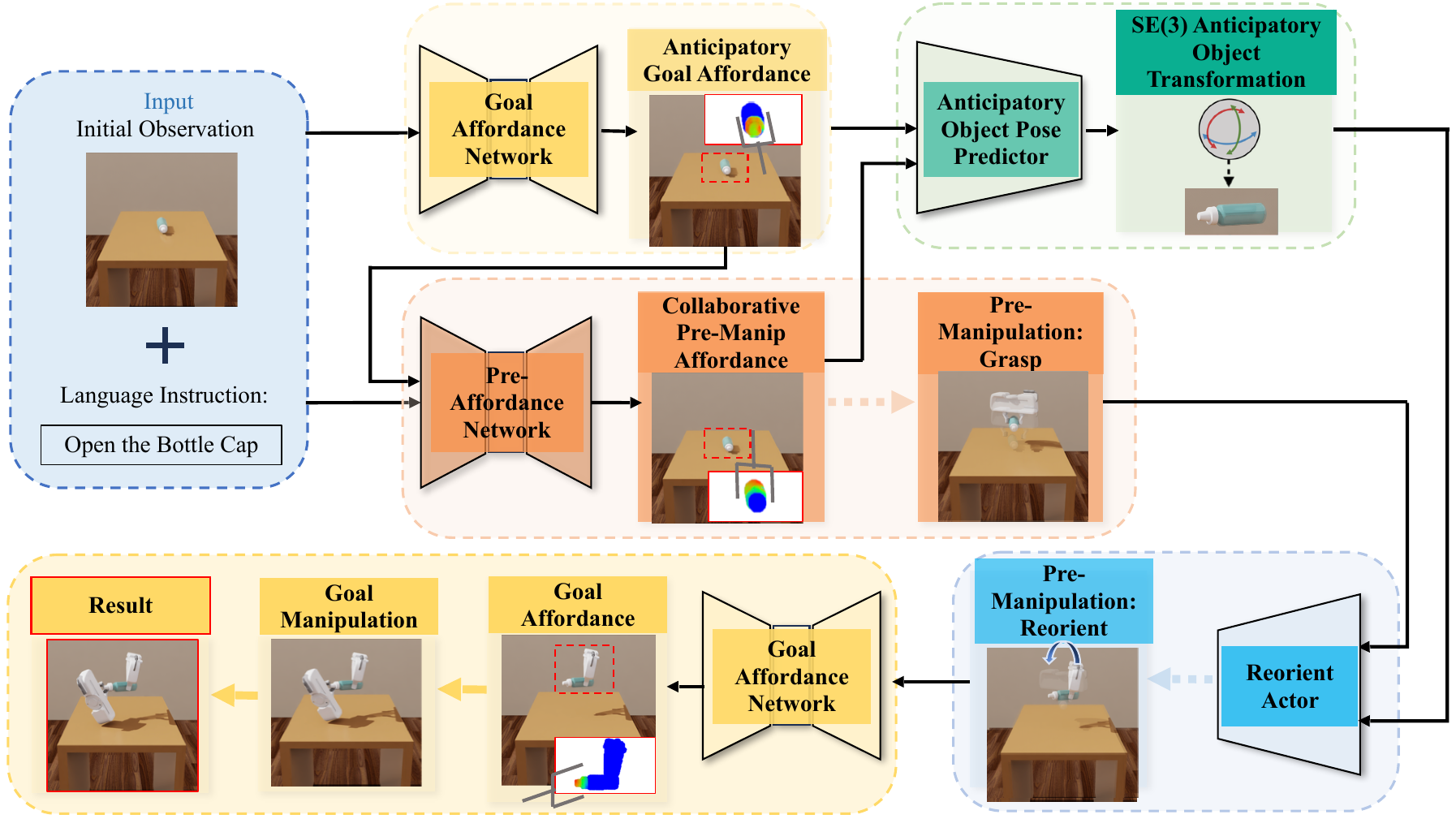}
\vspace{-6mm}
\caption{
\textbf{The BiPreManip pipeline.} Given the point cloud observation and a language instruction, the Goal Affordance Network predicts an anticipatory affordance for the primary arm. Conditioned on this, the Pre-Affordance Network infers how the assistant arm should act to establish favorable object conditions. The Anticipatory Object Pose Predictor and Reorient Actor estimate and execute the object reconfiguration required for collision-free access. Finally, the Goal Affordance Network is re-invoked on the updated scene to execute the goal-directed manipulation. This framework design enables anticipatory, collaborative, and geometrically consistent bimanual reasoning.
} 
\vspace{-3mm}
\label{fig:pipeline}
\end{figure*}

\subsection{Overview}

We introduce BiPreManip, a framework for bimanual preparatory manipulation, where one arm performs preparatory actions to enable the other’s goal-directed manipulation. As illustrated in Figure~\ref{fig:pipeline}, given an object point cloud and a language instruction, BiPreManip proceeds through a sequence of anticipatory reasoning and coordinated action prediction.
First, the Goal Affordance Network predicts an anticipatory goal affordance, envisioning the expected contact regions and action tendencies for the primary arm (\ref{subsec:goal_aff}). Conditioned on this prediction, the Pre-Affordance Network infers the collaborative pre-manipulation affordance, specifying how the assistant arm should act to pre-grasp the object while remaining coordinated with the primary arm’s future interaction (\ref{subsec:pre_aff}).
Based on these affordances, the Anticipatory Object Pose Predictor estimates an anticipatory object pose that enables collision-free access, and the Reorient Actor predicts the motion required to achieve it (\ref{subsec:reorient_network}). 
After these preparatory steps, the Goal Affordance Network is then re-invoked to produce the final affordance, guiding the primary arm to execute the goal-directed manipulation (\ref{subsec:goal_aff_2}).
The training objectives and loss formulations are detailed in Section~\ref{subsec:training_and_loss}.

\subsection{Goal Affordance Network} \label{subsec:goal_aff} 
Given an initial object point cloud $O \in \mathbb{R}^{N \times 3}$ and a language instruction $l$, the Goal Affordance Network models how the primary arm is expected to interact with the object to accomplish the specified goal. Concretely, it employs PointNet++~\cite{qi2017pointnet,qi2017pointnet++} to encode $O$ and extract per-point features $f_p, p \in {O}$, while the CLIP~\cite{radford2021learning} pretrained text encoder embeds the instruction $l$ into a textual feature $f_l$. These features are fused through a multilayer perceptron (MLP), which predicts a per-point score $s$ representing the likelihood that each point serves as a potential contact region. Aggregating these scores yields a dense affordance map that captures both spatial geometry and semantic relevance.

After identifying a high-score point $p_{\text{goal}}$, the network predicts the corresponding anticipatory action. A conditional variational autoencoder (cVAE)~\cite{bao2017cvae} is trained to model the distribution of goal-directed actions, conditioned on the concatenated features $(f_{p_{\text{goal}}}, f_l)$. The cVAE decoder outputs the gripper orientation $d_{\text{goal}} \in SO(3)$, which, combined with the contact point $p_{\text{goal}}$, defines a complete 6D action $a_{\text{goal}}=(p_{\text{goal}}, d_{\text{goal}}) \in SE(3)$.

Importantly, the predicted affordance map and the associated action are anticipatory rather than reactive, they envision interactions that are expected to be feasible after preparatory manipulation. 
This foresight allows the framework to reason about future object configurations and guide the assistant arm’s preparatory behavior accordingly.

\subsection{Pre-Affordance Network} \label{subsec:pre_aff}
Conditioned on the predicted anticipatory goal affordance, the Pre-Affordance Network infers the collaborative preparatory affordance and corresponding action for the assistant arm. These predictions are spatially and semantically aligned with the primary arm’s anticipated interaction.
The network integrates object geometry, language instruction, and the anticipated goal representation $(f_p, f_l, f_{p_{\text{goal}}}, f_{d_{\text{goal}}})$ through an MLP to estimate a per-point preparatory affordance map. A cVAE then samples the assistant gripper orientation $d_{\text{pre}}$ at the preparatory contact point $p_{\text{pre}}$, yielding the preparatory action $a_{\text{pre}} = (p_{\text{pre}}, d_{\text{pre}}) \in SE(3)$.
By explicitly reasoning about how the assistant arm can proactively establish conditions that facilitate the primary arm’s subsequent manipulation, the Pre-Affordance Network enables anticipatory, cooperative bimanual behavior and improves overall task efficiency.

\subsection{Object Pose Predictor and Reorient Actor}   \label{subsec:reorient_network}
In many preparatory manipulation tasks, the object’s pose must be adjusted before the primary arm can access the target region without interference from the assistant arm (\emph{e.g.}, rotating a bottle so that its lid faces the primary arm). To enable such reasoning, the Anticipatory Object Pose Predictor estimates a desired object pose that best facilitates the forthcoming goal-directed action, based on the predicted anticipatory goal action and the preparatory action. 
Specifically, the predictor takes as input the features of the goal action $(f_{p_{\text{goal}}}, f_{d_{\text{goal}}})$, the preparatory action $(f_{p_{\text{pre}}}, f_{d_{\text{pre}}})$, and the global object feature $f_O$, extracted from PointNet++. These features are fused and decoded to produce the desired object transformation $T^{\text{obj}} = (t^{\text{obj}}, r^{\text{obj}}) \in SE(3)$, where $t^{\text{obj}}$ and $r^{\text{obj}}$ denote the predicted translation and rotation, respectively. The transformation is applied to the object point cloud as
\begin{equation}
    O' = T^{\text{obj}} \cdot O,
\end{equation}
where each point $p$ in $O$ is transformed by $T^{\text{obj}}$. The resulting point cloud $O'$ represents the anticipated object configuration that provides unobstructed and geometrically favorable access for goal-directed manipulation.

Next, the Reorient Actor receives the predicted object configuration $O'$ together with the pre-grasped scene observation $O_{\text{grasped}}$, captured after the assistant arm has established stable contact.
PointNet++ encodes both inputs into global features $f_{O'}$ and $f_{O_{\text{grasped}}}$. These features are then used to predict the 6D preparatory motion required to reorient the object.
Both the predictor and actor are implemented as cVAEs. Together, they enable the model to envision and execute the object reconfiguration needed for effective bimanual coordination.

\subsection{Re-invoking the Goal Affordance Network} \label{subsec:goal_aff_2}
After preparatory manipulation, the Goal Affordance Network is re-applied to guide the primary arm’s final goal-directed action. Unlike the anticipatory stage, the input now reflects the updated scene containing the pre-manipulated object and the assistant gripper.

Using the same parameters as in the anticipatory stage, the network processes the updated observation and identical language instruction to predict the final affordance and corresponding action for goal execution. Sharing parameters across stages ensures semantic consistency between imagined and executed interactions, enabling BiPreManip to transition seamlessly from anticipatory reasoning to physical execution while preserving geometric coherence.

\begin{table*}[t!]
\caption{
Quantitative comparison of task success rates across different object categories and methods in simulation. For each category, results are reported on training (before slash) and unseen (after slash) objects.
}	
\vspace{-3mm}
\label{tab:result}	
	
\renewcommand{\arraystretch}{1.3}
\resizebox{\linewidth}{!}{
\begin{tabular}{c| c c c c c c c c c }
\toprule
\multirow{2}{*}{\textbf{}} &\multicolumn{9}{c}{\textbf {(1) Edge-Pushing}}\\
Method& Bowl & Cap & Keyboard & Laptop & Phone & Remote & Scissors & Switch & Window \\
\midrule
W2A & 
0\% / 0\% & 2\% / 4\% & 0\% / 0\% & 5\% / 10\% & 
2\% / 0\% & 0\% / 0\% & 5\% / 5\% & 0\% / 0\% & 
0\% / 0\% \\

ACT & 
32\% / 27\% & 22\% / 36\% & 2\% / 1\% & 7\% / 3\% & 
0\% / 0\% & 1\% / 0\% & 0\% / 0\% & 1\% / 1\% & 
1\% / 1\%  \\

3DA & 
0\% / 0\% & 0\% / 0\% & 0\% / 0\% & 1\% / 2\% & 
2\% / 0\% & 0\% / 0\% & 0\% / 0\% & 0\% / 0\% & 
1\% / 0\%  \\

3DFA & 
3\% / 0\% & 5\% / 14\% & 1\% / 3\% & 0\% / 2\% & 
1\% / 0\% & 0\% / 0\% & 0\% / 3\% & 0\% / 1\% & 
8\% / 9\%  \\

Heuristic & 
15\% / 21\% & 31\% / 37\% & 58\% / 62\% & 35\% / 33\% & 
30\% / 21\% & 20\% / 23\% & 15\% / 30\% & 20\% / 23\% & 
56\% / 56\% \\

\midrule
Ours & 
\textbf{49\%} / \textbf{52\%} & \textbf{71\%} / \textbf{74\%} & \textbf{64\%} / \textbf{64\%} & \textbf{62\%} / \textbf{67\%} & 
\textbf{66\%} / \textbf{42\%} & \textbf{31\%} / \textbf{24\%} & \textbf{34\%} / \textbf{49\%} & \textbf{61\%} / \textbf{72\%} & 
\textbf{87\%} / \textbf{87\%}  \\
\bottomrule

\multirow{2}{*}{\textbf{}}&\multicolumn{8}{c|}{\textbf {(2) Articulated Manipulation} }&\multicolumn{1}{c}{\textbf {(3)}}\\
			
Method
& Bottle & Dispenser & Lighter & Pen-Button & Pen-Cap & Pliers & Stapler & \multicolumn{1}{c|}{USB} & Plate\\
\midrule
W2A & 
1\% / 2\% & 0\% / 1\% & 2\% / 0\% & 0\% / 0\% & 
1\% / 0\% & 9\% / 8\% & 16\% / 16\% & \multicolumn{1}{c|}{5\% / 3\%} & 
4\% / 4\% \\

ACT & 
2\% / 0\% & 54\% / 43\% & 34\% / 30\% & 15\% / 9\% & 
0\% / 0\% & 24\% / 8\% & \textbf{38\%} / 23\% & \multicolumn{1}{c|}{1\% / 1\%} & 
30\% / 26\% \\

3DA & 
0\% / 0\% & 20\% / 2\% & 20\% / 3\% & 1\% / 1\% & 
0\% / 0\% & 1\% / 7\% & 6\% / 3\% & \multicolumn{1}{c|}{0\% / 1\%} & 
27\% / 25\% \\

3DFA & 
4\% / 2\% & \textbf{57\%} / 41\% & 41\% / 36\% & 14\% / 25\% & 
1\% / 0\% & 2\% / 5\% & 8\% / 1\% & \multicolumn{1}{c|}{12\% / 0\%} & 
71\% / 68\% \\

Heuristic & 
19\% / 14\% & 31\% / 47\% & 21\% / 32\% & 27\% / 34\% & 
22\% / 15\% & 6\% / 10\% & 19\% / 21\% & \multicolumn{1}{c|}{12\% / 10\%} & 
81\% / 78\% \\

\midrule
Ours & 
\textbf{30\%} / \textbf{26\%} & 45\% / \textbf{56\%} & \textbf{43\%} / \textbf{58\%} & \textbf{67\%} / \textbf{72\%} & 
\textbf{26\%} / \textbf{32\%} & \textbf{25}\% / \textbf{29}\% & \textbf{38\%} / \textbf{30\%} & \multicolumn{1}{c|}{\textbf{13\%} / \textbf{14\%}} & 
\textbf{85\%} / \textbf{82\%} \\
\bottomrule

\end{tabular}
}

\small
\vspace{-1mm}
\end{table*}

\subsection{Training and Losses} \label{subsec:training_and_loss}
This section describes the training objectives and loss functions used in the BiPreManip framework.

\subsubsection{Affordance Network Loss}
\paragraph{Affordance Score.}
Following~\cite{mo2021where2act, zhao2022dualafford}, affordance prediction is supervised using both successful and failed demonstrations. A critic network estimates the empirical success probability of sampled actions over the object surface, forming the ground-truth affordance map.
The predicted per-point affordance scores are optimized via an $\ell_1$ loss:
\begin{equation}
\mathcal{L}_{\mathrm{aff}} = \|s_{\text{pred}} - s_{\text{gt}}\|_1,
\end{equation}
where $s_{\text{pred}}$ and $s_{\text{gt}}$ denote the predicted and ground-truth affordance scores, respectively.

\paragraph{Gripper Orientation.}
For the cVAE modules that predict 3D gripper orientations $d \in SO(3)$, we employ a combination of geodesic distance loss and KL divergence regularization.
The geodesic loss measures the angular discrepancy between the predicted and ground-truth orientations:
\begin{equation}
\mathcal{L}_{\mathrm{ori}}(d, d^*) = \arccos\!\left(\frac{\operatorname{Tr}\!\left(d^{\top} d^*\right) - 1}{2}\right),
\label{eq:geodesic_distance}
\end{equation}
where $d^*$ is the ground-truth orientation obtained from demonstrations. The latent distribution is regularized using:
\begin{equation}
\mathcal{L}_{\mathrm{KL}} = 
D_{\mathrm{KL}}\!\left(
q_\phi(z \mid d^*, c)\,\|\, \mathcal{N}(0,1)
\right),
\label{eq:kl_general}
\end{equation}
where $c$ denotes the conditioning context.

\vspace{-3mm}
\paragraph{Supervision in the Anticipatory Stage.}
Unlike the preparatory and execution stages, the anticipatory stage lacks explicit annotations since it does not directly appear in demonstrations. To provide supervision, we align the anticipatory-stage object with the execution-stage object to construct ground truth.

Specifically, for the anticipatory affordance, ground-truth affordance scores from the execution stage are mapped to the anticipatory frame using the known object pose transformation (obtained from simulation or pose-estimation pipelines~\cite{wen2024foundationpose,thalhammer2024challenges}). Each execution-stage point’s success probability label is reused for its corresponding anticipatory point, ensuring that desired contact regions remain consistent before and after preparatory manipulation.

For the anticipatory gripper orientation, ground truth is similarly derived by transforming the final-stage gripper orientation according to the object’s pose change.
Given the object’s initial and final rotations $R_{\text{obj,init}}$ and $R_{\text{obj,fin}}$, and the final-stage gripper orientation $R_{\text{grp,fin}}$, the anticipatory ground-truth is obtained as:
\begin{equation}
R_{\text{grp,ant}} = R_{\text{obj,init}} \cdot R_{\text{obj,fin}}^{\top} \cdot R_{\text{grp,fin}}.
\end{equation}
This formulation assumes that the relative transformation between the gripper and the object remains constant across stages, ensuring geometric consistency between the anticipated and executed interactions.

\subsubsection{Losses for Pose Predictor and Reorient Actor}
Both the Anticipatory Object Pose Predictor and the Reorient Actor are trained as cVAEs that output a 6D object transformation $M = (r, t) \in SE(3)$.
The rotation $r$ is supervised using the geodesic loss, the translation $t$ with an $\ell_1$ loss, and the latent distribution with a KL divergence term. The overall objective is:
\begin{equation}
\mathcal{L}_{\mathrm{M}} = 
\mathcal{L}_{\mathrm{ori}}(r, r^*) + 
\mathcal{L}_{\mathrm{1}}(t, t^*) + 
\mathcal{D}_{\mathrm{KL}}.
\end{equation}

\subsubsection{Overall Optimization}
The total training loss is a weighted sum of the above components.
Each module is trained with its respective objectives, while gradients are propagated through the shared feature encoders to maintain consistent feature representations across networks.
This modular optimization strategy stabilizes learning while preserving coherent coordination across anticipatory, preparatory, and execution stages.

\setlength{\parindent}{0pt}
\vspace{-1mm}
\section{Experiments}
\label{sec:exp}

\vspace{-1mm}
\subsection{Simulation and Settings}
\vspace{-1mm}
\label{subsec:setup}
The simulation environment is constructed on the SAPIEN~\cite{Xiang_2020_SAPIEN} platform, utilizing the Franka Panda grippers as the robotic end-effector. 
To comprehensively evaluate the model’s manipulation capability, we design three task types:
(1) Articulated Manipulation Tasks, where articulated objects must be lifted, rotated, or repositioned to enable functional manipulation of their movable parts;
(2) Edge-Pushing Tasks, involving flat or weakly graspable objects that must be pushed toward the table edge before grasping; and
(3) Plate-Lifting Tasks, adapted from the PerAct2 Benchmark~\cite{grotz2024peract2}, where one arm presses down one edge of a plate while the other grasps the lifted side.

From the ShapeNet~\cite{chang2015shapenet} and PartNet-Mobility~\cite{mo2019partnet, Xiang_2020_SAPIEN} datasets, we select 18 object categories comprising 882 instances, emphasizing objects that are challenging to directly grasp or manipulate functionally. Objects are randomly divided into training and unseen (novel) sets with a 3:1 ratio. 
During data collection, an object from a random category is placed near the table center with a random 6D pose. A camera positioned above the scene with a downward viewing angle captures the partial 3D point cloud. For each category, we collect 1,000 successful demonstrations and 1,000 failure cases.
For evaluation, we construct two inference sets: one containing seen (training) objects with randomly varied initial poses, and another containing unseen (novel) objects with distinct shapes and poses. Each set includes 100 test samples per category. Additional dataset details are provided in the supplementary material.

\vspace{-3mm}
\paragraph{Evaluation Metrics.} 
We use task-specific success criteria for quantitative evaluation:
(1) Articulated Manipulation Tasks are considered successful if the functional part (\emph{e.g.}, a bottle lid or dispenser pump head) is correctly operated (\emph{e.g.}, opened or pressed);
(2) Edge-Pushing Tasks succeed if the primary gripper successfully grasps and lifts the object; and
(3) Plate-Lifting Tasks succeed if the plate is securely grasped and lifted by the primary gripper.

\begin{figure*}[t]
\centering
\includegraphics[width=1\linewidth]{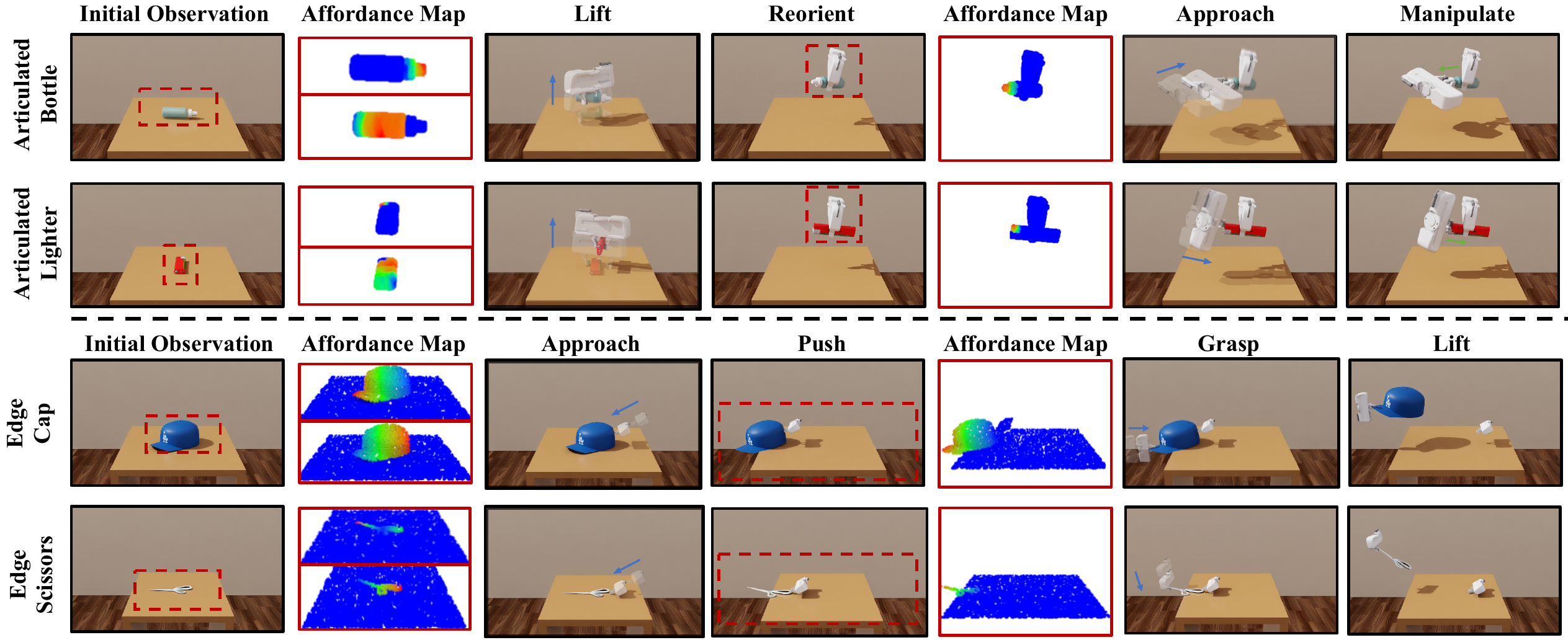}
\vspace{-6mm}
\caption{
\textbf{Simulation experiments.} Qualitative results of the predicted affordance maps and corresponding actions from our method. In each example, the second column presents the predicted affordances: the top subfigure depicts the \textit{anticipatory affordance} for the primary arm, while the bottom subfigure shows the \textit{pre-manipulation affordance} for the assistant arm.
} 
\vspace{-2mm}
\label{fig:sim}
\end{figure*}

\subsection{Baselines and Ablations}
\vspace{-1mm}

We compare our method with representative baselines spanning three major paradigms in robot manipulation: geometry-aware affordance methods that infer actions through explicit 3D reasoning, imitation learning approaches that directly predict bimanual actions from demonstrations, and rule-based strategies based on hand-crafted heuristics. 
Together, these baselines cover a diverse set of modeling assumptions and highlight the challenges of collaborative preparatory manipulation. 
All learning-based baselines are trained from scratch on identical rollout trajectories under a multi-category setting, ensuring consistent data distribution and a fair comparison across methods.

\begin{itemize}
    \item \textbf{W2A}~\cite{mo2021where2act}: 
    An affordance-based method that predicts goal-directed actions from object geometry and generalizes well across categories. However, as an inherently single-arm approach, it models only goal affordances and does not consider bimanual coordination or preparatory manipulation, and thus also serves as a reference for assessing the role of coordination in our tasks.
    \item \textbf{ACT}~\cite{zhao2023learning}: This method employs a transformer-based encoder-decoder architecture with an action-chunking strategy to predict bimanual action sequences.
    \item \textbf{3DA}~\cite{ke20243d}: A diffusion-policy framework that learns 3D action trajectories from point cloud observations, showing strong spatial reasoning and motion generation. We adapt it to a dual-arm setting by extending the action space for coordinated control.
    \item\textbf{3DFA}~\cite{gkanatsios20253d}: A flow-matching variant of 3DA that improves training stability and trajectory precision. This method represents the current state of the art in bimanual manipulation learning.  
    \item\textbf{Heuristic}: A rule-based baseline that employs hand-crafted strategies with access to ground-truth scene information in simulation. Implementation details are provided in the supplementary material. 
    \item\textbf{w/o Ant-Aff}: An ablation removing the Goal Affordance Network in the anticipatory stage. 
    \item\textbf{w/o ObjPosePred}: An ablation removing the Anticipatory Object Pose Predictor module.
\end{itemize}

\begin{table*}[htbp]
\centering
\caption{
Ablation study results on the Articulated Manipulation Tasks.
}
\vspace{-3mm}
\label{tab:ablation}
\small
\renewcommand{\arraystretch}{1.3}
\resizebox{\linewidth}{!}{
\begin{tabular}{c c c c c c c c c c }
\toprule
Method & Bottle & Dispenser & Lighter & Pen-button & Pen-cap & Pliers & Stapler & USB \\
\midrule

w/o Ant-Aff & 
27\% / 13\% & 39\% / 48\% & 39\% / 48\% & 48\% / 58\% & 
23\% / 10\% & 8\% / 12\% & 21\% / 23\% & 3\% / 3\% \\

w/o ObjPosePred & 
24\% / 15\% & 38\% / 45\% & 31\% / 52\% & 51\% / 50\% & 
21\% / 8\% & 20\% / 31\% & 14\% / 14\% & 10\% / 6\%  \\
\midrule
ours & 
\textbf{30\%} / \textbf{26\%} & \textbf{45}\% / \textbf{56\%} & \textbf{43\%} / \textbf{58\%} & \textbf{67\%} / \textbf{72\%} & 
\textbf{26\%} / \textbf{32\%} & \textbf{25\%} / \textbf{29\%} & \textbf{38\%} / \textbf{30\%} & \textbf{13\%} / \textbf{14\%} \\
\bottomrule
\end{tabular}
}

\end{table*}

\begin{figure*}[t]
\centering
\includegraphics[width=1\linewidth]{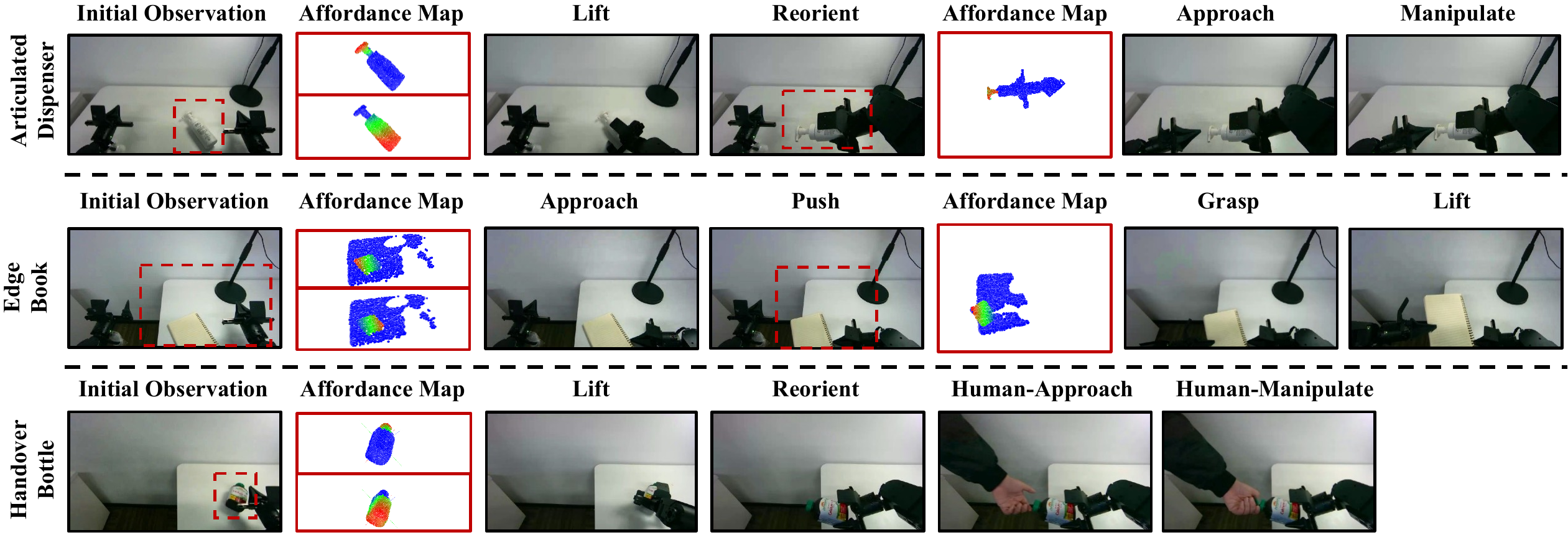}
\vspace{-7mm}
\caption{
\textbf{Real-world results} across different objects and task types.
} 
\label{fig:real}
\end{figure*}

\vspace{-1mm}
\subsection{Results and Analysis} \label{subsec:analysis}
Table~\ref{tab:result} reports the success rate comparisons across different methods on both seen and unseen objects in the simulation. Our method outperforms all baselines, highlighting its effectiveness and geometric generalization capability.

The \textbf{W2A} baseline yields relatively low success rates across all task categories, despite its strong performance on geometry-aware manipulation in prior work. This indicates that many objects cannot be directly grasped or operated on without preparatory actions, underscoring the necessity of collaborative bimanual pre-manipulation.

For \textbf{ACT}, although the model effectively learns temporal action patterns from demonstrations, its predictions are primarily conditioned on RGB images and joint states. As a result, the predicted actions may appear reasonable in the 2D image space but do not always correspond to physically feasible 3D interactions, sometimes leading to misaligned grasps (\emph{e.g.}, approaching beside a bottle cap).

For \textbf{3DA}, the predicted actions tend to be spatially coarse, often leading to suboptimal grasp configurations. The diversity of object geometries further increases task complexity, as the model frequently struggles to infer the precise approach height required for reliable contact. Even small vertical deviations can cause the gripper to miss the object or collide with the table surface.

\textbf{3DFA} improves upon 3DA through flow matching, yet its generated actions are not consistently aligned with the geometric constraints or functional requirements of a task. In articulated manipulation, for instance, it may pre-grasp near a bottle cap rather than the midsection, restricting the available workspace for the other arm and increasing collision risk. This reflects the absence of anticipatory coordination, which is crucial for successful bimanual execution.

The \textbf{Heuristic} baseline achieves relatively high scores in some tasks, however, it relies on predefined canonical poses and accurate object pose information—assumptions that rarely hold in real-world settings. Consequently, its generalization and practicality are limited.

In contrast, \textbf{our method} effectively learns to perform bimanual preparatory manipulation across diverse object categories, as presented in Figure~\ref{fig:sim}.
For articulated objects (top two rows), the assistant arm first lifts and reorients the object to expose the functional part, enabling the primary arm to complete the manipulation guided by the predicted affordance map.
For flat or weakly graspable objects (bottom two rows), 
the assistant arm first approaches a contact point that maintains object stability during pushing, then repositions it into a configuration suitable for grasping, enabling the primary arm to successfully grasp and lift an object that was initially inaccessible.
This progression from anticipation to cooperation demonstrates the model’s ability to reason over long-horizon dependencies between preparatory and goal-directed actions. Additional visualizations are provided in the supplementary material.

We conduct ablation studies on the most representative and challenging tasks, the Articulated Manipulation Tasks, and the results in Table~\ref{tab:ablation} highlight the contribution of each component.
Removing the Goal Affordance Network in the anticipatory stage (\textbf{w/o Ant-Aff}) yields a notable performance drop, indicating that learning goal-directed affordances is crucial for guiding effective preparatory behaviors.
Similarly, omitting the Object Pose Predictor (\textbf{w/o ObjPosePred}) degrades performance, confirming that envisioning feasible target poses enables the assistant arm to reposition objects more accurately.
Together, these findings underscore the importance of anticipatory reasoning and coordinated bimanual manipulation for achieving robust preparatory behavior.

\begin{table}[htbp]
\centering
\footnotesize
\caption{\textbf{Real-world success rate} in different tasks.}
\vspace{-2mm}
\label{tab:real_exp}


\renewcommand{\arraystretch}{1.2} 
\resizebox{\linewidth}{!}{
\begin{tabular}{@{\extracolsep{\fill}} l|c c|c c|c c@{}} 
\toprule
 & \multicolumn{2}{c|}{\textbf{Edge}} & \multicolumn{2}{c|}{\textbf{Articulated}} & \multicolumn{2}{c}{\textbf{Handover}}  \\ 
 Method & Book \ & Hat & Bottle & Dispenser & Bowl & Bottle  \\
\midrule
W2A & 
0/10 & 1/10 & 0/10 & 1/10 & 
0/10 & 0/10 \\
3DFA & 
1/10 & 0/10 & 0/10 & 2/10 & 
0/10 & 2/10 \\
\midrule
ours & 
7/10 & 8/10 & 6/10 & 8/10 & 
5/10 & 6/10 \\
\bottomrule
\end{tabular}
}

\end{table}

\subsection{Real-World Evaluation}
We conduct real-world experiments on an ARX-X7s dual-arm platform, using an Intel RealSense L515 depth camera to capture partial 3D point clouds as input to our model. Figure~\ref{fig:real} shows qualitative results on various real-world objects, while Table~\ref{tab:real_exp} reports the quantitative comparisons of success rates against baseline methods. 
We further evaluate a robot–human handover scenario, where the robot leverages the learned policy to anticipate the human’s intended goal and perform preparatory actions to deliver the object in a configuration that facilitates human use. Additional results are provided in the supplementary material and video.

\vspace{1mm}
\section{Conclusion}
\label{sec:conclusion}

In this paper, we introduce Collaborative Preparatory Manipulation, a class of bimanual tasks where one arm performs preparatory manipulations to enable the other’s goal-directed action. To tackle these asymmetric and long-horizon tasks, we propose BiPreManip, a visual affordance-based framework that anticipates the final interaction and guides preparatory behavior through anticipatory affordance reasoning and object pose reconfiguration. Extensive experiments in both simulation and the real world demonstrate that BiPreManip substantially outperforms existing baselines, underscoring the importance of anticipatory coordination for robust and intelligent dual-arm manipulation.
\newpage
{
    \small
    \bibliographystyle{ieeenat_fullname}
    \bibliography{main}
}





\clearpage
\appendix
\maketitlesupplementary

\begin{strip}
    \centering
    \vspace{2mm}
    \includegraphics[width=1\textwidth]{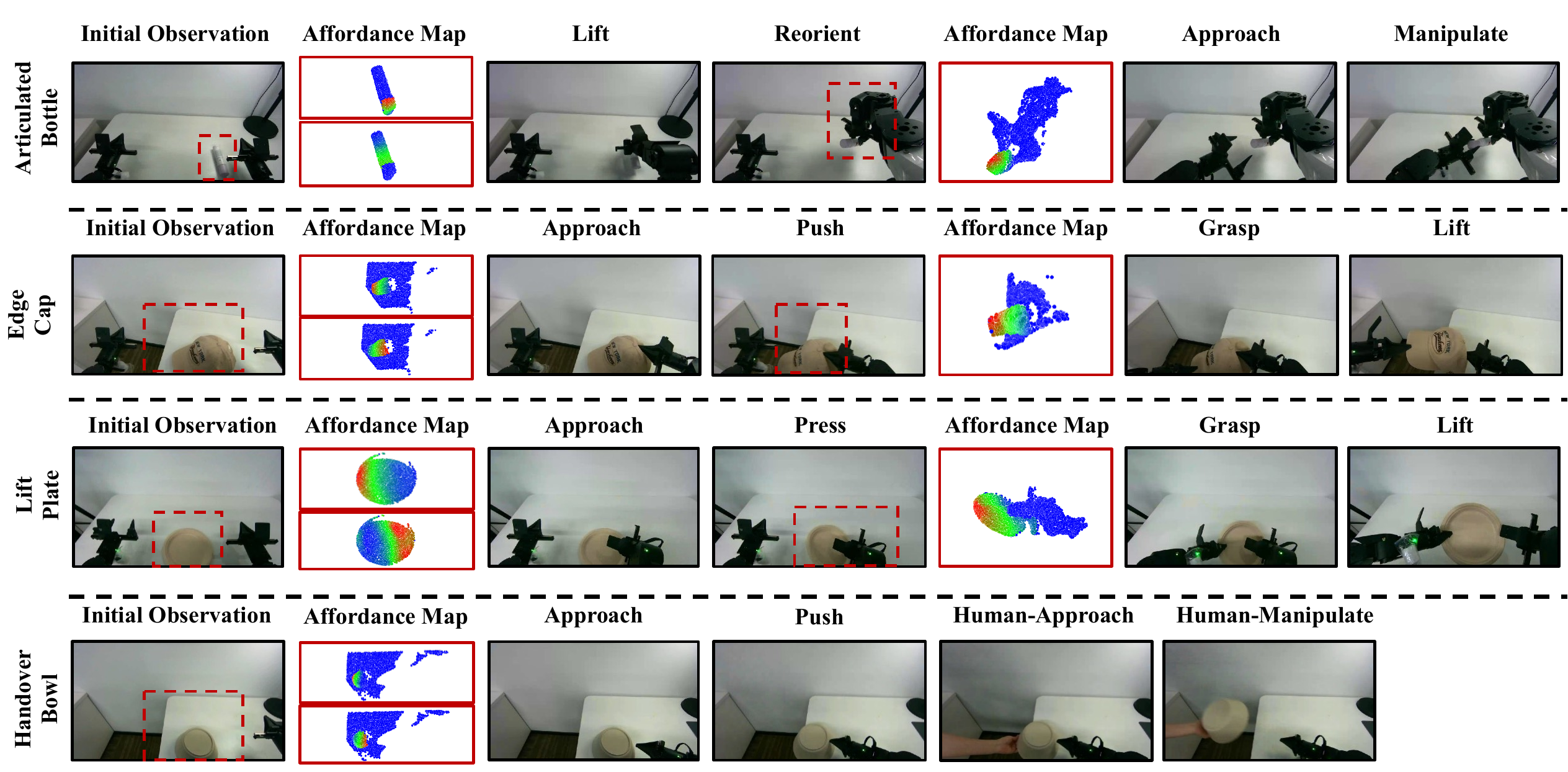}
    \captionof{figure}{
\textbf{Real-World experiments.} We present additional qualitative visualizations of the predicted affordance maps and corresponding actions generated by our method. In
each example, the second column displays the predicted affordances: the top subfigure depicts the \textit{anticipatory affordance} for the primary arm, while the bottom subfigure shows the \textit{pre-manipulation affordance} for the assistant arm. Manipulation videos are included in the supplementary material.
}  
    \label{fig:supp_real}
    \vspace{2mm}
\end{strip}

\section{Additional Real-World Results}\label{supp:real_exp}

In real-world experiments, the predicted $\mathrm{SE}(3)$ gripper poses are first validated for inverse kinematics feasibility and controller constraints, and then executed via time-parameterized interpolation with linear translation and SLERP-based rotation under predefined execution tolerances.
Actions that fail these feasibility checks are recorded as failures without recovery or resampling (but deployment users may instead resample actions from the affordance and retry execution).

Figure~\ref{fig:supp_real} presents additional real-world qualitative results that complement the visualizations shown in Figure~5 of the main paper.

For each example, we provide the predicted affordance maps and the corresponding actions generated by our method. In the second column, the top subfigure illustrates the \textit{anticipatory affordance} for the primary arm, while the bottom subfigure depicts the \textit{pre-manipulation affordance} for the assistant arm. The fourth column displays the \textit{goal affordance} associated with the primary arm’s goal-directed action. Together, these visualizations highlight how the model effectively coordinates both arms to achieve stable pre-manipulation behaviors that enable successful task execution.

Beyond the articulated manipulation, edge-pushing, and plate-lifting tasks, we further evaluate a robot–human handover scenario. This experiment demonstrates that our approach is not only effective for bimanual collaborative preparatory manipulation but also readily extends to robot–human interaction. 
In this new setting, the model anticipates the human’s intended goal, analogous to how it predicts the primary arm’s objective in the bimanual manipulation scenario, and acts as an assistant by performing preparatory actions that deliver the object in a configuration suitable for comfortable and efficient human use.
This scenario further illustrates the versatility and practical applicability of our proposed method.

Manipulation videos are provided in the supplementary material.

\section{Additional Visualizations in Simulation}
\label{supp:sim_visu}

\begin{figure*}[t]
\centering
\includegraphics[width=1\linewidth]{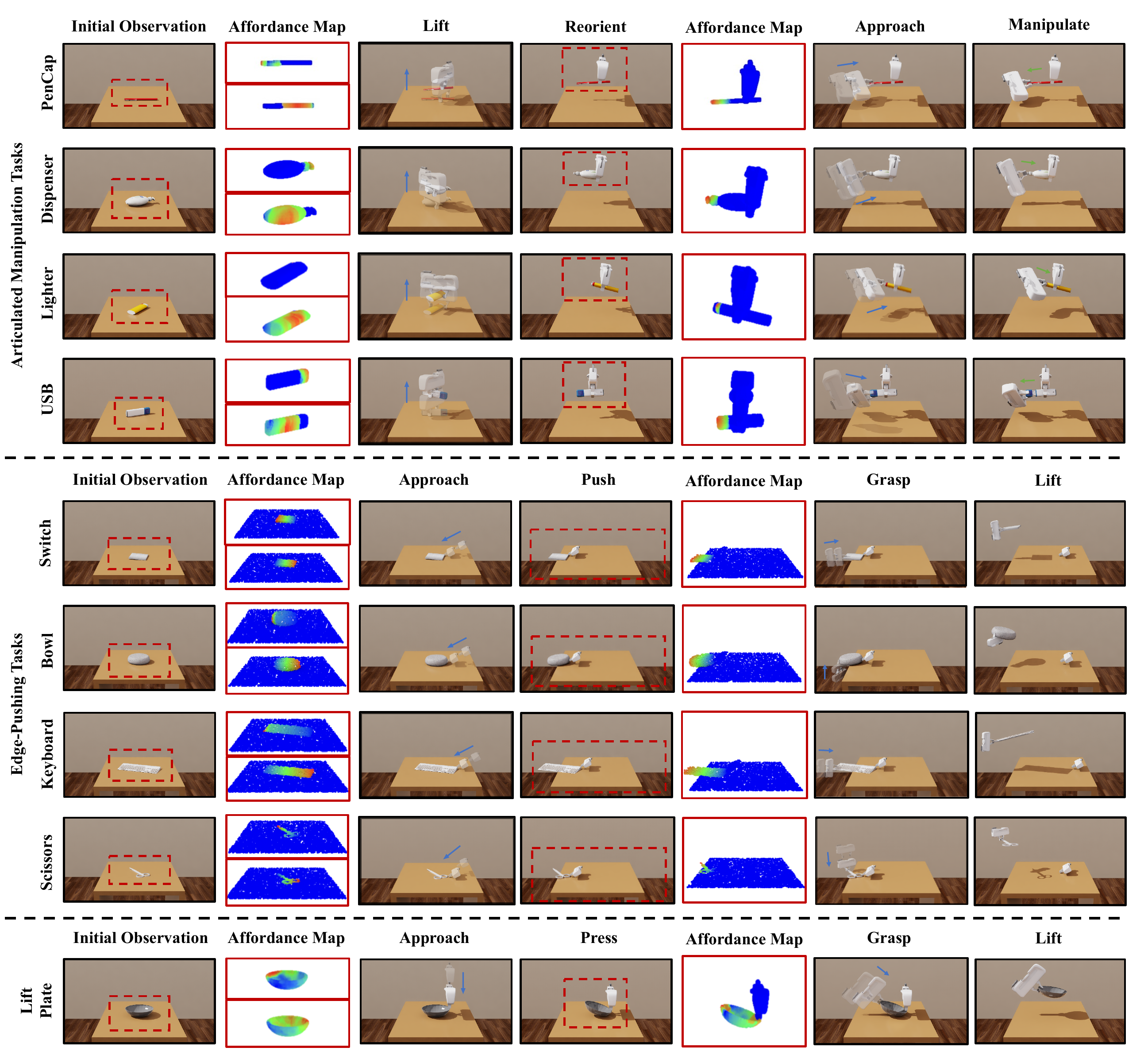}
\caption{
\textbf{Simulation experiments.} We present additional qualitative visualizations of the predicted affordance maps and corresponding actions produced by our method. In
each example, the second column displays the predicted affordances: the top subfigure depicts the \textit{anticipatory affordance} for the primary arm, while the bottom subfigure shows the \textit{pre-manipulation affordance} for the assistant arm.
} 
\label{fig:supp_sim}
\end{figure*}

We present additional visualization results from simulation in Figure~\ref{fig:supp_sim}, covering different task types and object categories, to augment Figure 4 in the main paper.
\section{Data Statistics} \label{supp:data}

We provide detailed statistics of the datasets used in our experiments in Table~\ref{supptab:dataset_stats}. This table lists the number of object instances in the training and unseen test sets for each object category across all task types. As described in Section~4.1 of the main paper, objects are randomly divided into training and unseen (novel) sets with a 3:1 ratio. After training the model on the training objects, we construct two evaluation sets: one containing seen (training) objects with randomly varied initial poses, and another containing unseen (novel) objects with distinct shapes and poses.

We further note that, due to the object-level random split, some unseen (novel) objects may exhibit simpler geometries than those in the training set, leading to consistently higher accuracy across all methods for certain categories.

\begin{table}[htbp]
\centering
\small
\caption{Number of object instances per category in the training and unseen test sets.}
\label{supptab:dataset_stats}
\setlength{\tabcolsep}{7pt}
\renewcommand{\arraystretch}{1.25}
\resizebox{\linewidth}{!}{
\begin{tabular}{p{2.8cm} | l c c c}
\toprule
\textbf{Task Type} & \textbf{Category} & \textbf{Train} & \textbf{Unseen} & \textbf{Total} \\
\midrule

\multirow{9}{=}{Edge-Pushing Tasks} 
& Bowl      & 140 & 46 & 186 \\
& Cap       & 42  & 14 & 56 \\
& Keyboard  & 49  & 16 & 65 \\
& Laptop    & 42  & 13 & 55 \\
& Phone     & 14  & 4  & 18 \\
& Remote    & 37  & 12 & 49 \\
& Scissors  & 36  & 11 & 47 \\
& Switch    & 53  & 17 & 70 \\
& Window    & 44  & 14 & 58 \\
& \textbf{Subtotal} & \textbf{457} & \textbf{147} & \textbf{604} \\

\midrule
\multirow{7}{=}{Articulated\\Manipulation Tasks}
& Bottle    & 43  & 14 & 57 \\
& Dispenser & 43  & 14 & 57 \\
& Lighter   & 12  & 3  & 15 \\
& Pen       & 36  & 12 & 48 \\
& Pliers    & 19  & 6  & 25 \\
& Stapler   & 18  & 5  & 23 \\
& USB       & 23  & 7  & 30 \\
& \textbf{Subtotal} & \textbf{194} & \textbf{61} & \textbf{255} \\

\midrule
\multirow{2}{=}{Plate-Lifting Tasks}
& Plate     & 17  & 5  & 22 \\
& \textbf{Subtotal} & \textbf{17} & \textbf{5} & \textbf{22} \\

\midrule
\rowcolor{gray!20}
\textbf{All Tasks} & \textbf{Overall Total} & \textbf{668} & \textbf{213} & \textbf{882} \\
\bottomrule
\end{tabular}
}
\end{table}

\section{Dataset Collection and Heuristic Baseline} \label{supp:data_collection}

This section details the data collection procedures for the three simulated task sets: (1) Articulated Manipulation, (2) Edge-Pushing, and (3) Plate-Lifting. To enable efficient large-scale data generation, we incorporate a set of hand-crafted heuristic strategies that guide action selection and substantially reduce failed rollouts. 

The Heuristic Baseline reported in the main paper is implemented using the same set of heuristics.

\subsection{Articulated Manipulation Tasks}

For each trial, we randomly sample an articulated object from the PartNet-Mobility dataset~\cite{mo2019partnet, Xiang_2020_SAPIEN}. The object is placed near the table center with a randomized 6D pose, and an overhead depth camera provides RGB-D observations throughout the episode.

\subsubsection{Pre-Grasp Action}
The first step is to generate a pre-grasp pose for the assistant arm. Using the depth map and link masks rendered from simulation, we sample a pixel on a non-articulated (fixed) link of the object. This bias toward the non-articulated part helps reduce the likelihood of the assistant arm interacting with articulated parts, thereby mitigating potential inter-arm interference during the subsequent goal-directed manipulation. The selected pixel is then back-projected to obtain a 3D grasp point.

The gripper’s orientation is determined using simple geometric rules. The gripper’s forward direction (x-axis) is oriented approximately downward toward the object, with a small (10°) randomized deviation for diversity. The left direction (y-axis) is defined according to a category-specific canonical direction in the object’s local coordinate frame (e.g., for elongated objects, the left direction is set perpendicular to the object’s principal axis). This direction is then transformed from the object’s local coordinate frame to the world frame, based on the object’s sampled pose, promoting consistency of the gripper orientation relative to the object across different placements. A small random angular perturbation is added for further diversity.

Together, the forward (x), left (y), and derived up (z) axes define a complete 6-DoF pre-grasp pose, which is used for the assistant arm to grasp. 
The assistant arm then lifts the object by elevating the gripper.

\subsubsection{Reorientation Action}

After grasping, the assistant arm reorients the object to expose its articulated component to the primary arm. The desired object pose is defined using category-dependent rules. For example, in the case of a bottle, the lid is rotated to face the primary arm and positioned within the shared reachable workspace of both arms.

Let $T_{\text{obj}}^{\text{grasped}} \in SE(3)$ and $T_{\text{grp}}^{\text{grasped}} \in SE(3)$ denote the object and gripper poses after grasping. The desired object pose,  $T_{\text{obj}}^{\text{reorient}}$, defines the target gripper pose $T_{\text{grp}}^{\text{reorient}}$ by preserving the relative transform between the gripper and the object:
\begin{equation}
    T_{\text{grp}}^{\text{reorient}}
    = T_{\text{obj}}^{\text{reorient}}
    \left(T_{\text{obj}}^{\text{grasped}}\right)^{-1}
    T_{\text{grp}}^{\text{grasped}}.
\end{equation}

The assistant arm then executes this reorientation action, positioning the object into the desired configuration to enable reliable goal-directed manipulation by the primary arm.

\subsubsection{Goal-Directed Action}
A new depth image is rendered after reorientation. We then sample a pixel on the articulated link and back-project it to obtain the target 3D contact point. The gripper’s forward (x) axis is aligned with the articulation’s functional direction (e.g., opening or pushing), with a small probability. The left (y) and up (z) axes are randomly assigned to introduce variability.

\subsection{Edge-Pushing Tasks}
For each trial, an object is randomly sampled from the ShapeNet~\cite{chang2015shapenet} or PartNet-Mobility~\cite{mo2019partnet, Xiang_2020_SAPIEN} datasets. The object is then placed on the table with a randomized pose.

The assistant arm begins with preparation manipulation. A random pixel is sampled from the object in the depth image and back-projected to determine the 3D contact point. The gripper's forward (x) axis is oriented roughly downward toward the object, with a random deviation for variability. The gripper’s left (y) axis is roughly aligned with the direction of the table edge toward which the object is intended to be pushed, biasing the configuration so that the gripper’s fingers, rather than the gripper body, are more likely to contact the object. Together, this orientation and the selected contact point define the gripper’s initial approach pose. The robot then pushes the object toward the table edge, while maintaining the orientation and translating the gripper in the pushing direction. Random deviations are applied to introduce variability in the poses.

After the preparatory motion, if the object remains on the table, the primary arm attempts to grasp it at the edge. A new depth image is captured, and a pixel is sampled from the part of the object extending beyond the table, from which a new 3D contact point is computed. The gripper’s orientation is then determined based on the object’s geometry. For thin objects, such as a keyboard or phone, the gripper’s forward (x) axis is aligned horizontally and perpendicular to the exposed object edge, while the left (y) axis is aligned vertically relative to the table surface. For objects like an inverted bowl, the gripper’s forward (x) axis is oriented from underneath to grasp the bowl’s rim, with the left (y) axis assigned randomly. All generated axes are perturbed with small random deviations to increase pose diversity.
Finally, after the primary arm grasps the object, it lifts the object to a higher position. The grasp is considered successful if the object remains stable and does not fall, indicating that the primary arm has maintained a secure grip.

\subsection{Plate-Lifting Tasks}

This task is adapted from the PerAct2~\cite{grotz2024peract2} benchmark. Following their setup, the assistant arm first presses down one side of the plate. To do this, we randomly select a point along the plate’s outer edge as the pressing point. The gripper’s forward (x) axis is oriented vertically downward, and its left (y) axis is aligned with the tangent direction of the plate’s boundary at the selected point. Small random perturbations are applied to each axis to introduce orientation variability.
Once the plate is pressed down, the primary arm selects a grasping point from the top 10\% of points on the plate with the highest z-coordinates. For the chosen point, the gripper’s forward (x) axis is aligned normal to the plate edge at that location, while the up (z) axis is aligned parallel to the corresponding tangent direction, promoting a stable grasp. After closing the gripper, the primary arm lifts the plate to a higher position. The grasp is considered successful if the plate remains stable and does not fall.

\section{Training Details and Computational Costs} 
\label{supp:cost}

As described in Section 3.6.3 of the main paper, all modules in our framework are jointly optimized using a weighted sum of their respective loss functions. Gradients are propagated through shared feature encoders, promoting consistent and coherent feature representations across the network.

Training the BiPreManip framework on a single NVIDIA V100 GPU requires approximately 24 hours to converge.

During inference, the framework consumes 1,166 MB of GPU memory. The average model inference time (excluding physics simulation) for a complete rollout is 0.27 seconds. Specifically, the pre-grasping manipulation stage takes approximately 0.12 seconds, the reorientation stage requires 0.08 seconds, and the final goal-directed action stage takes 0.07 seconds.

\section{Additional Experiment on Occlusion} 
\label{supp:occlusion}

\begin{figure}[htbp]
\centering
\includegraphics[width=\linewidth]{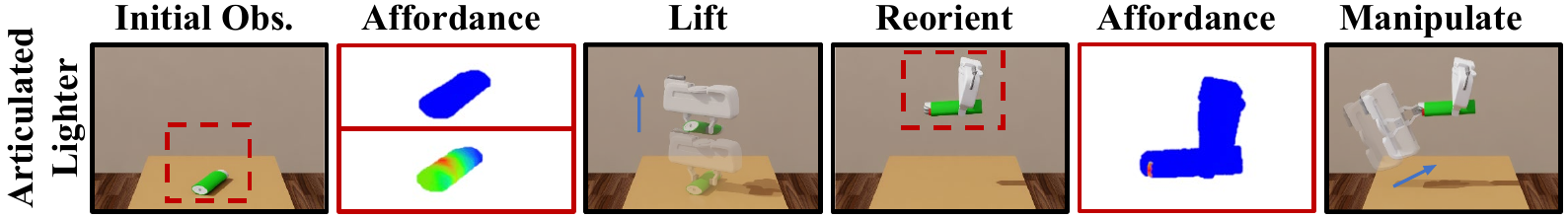}
\caption{
An example where the functional part is occluded in the initial observation.
} 
\label{fig:occlusion}
\end{figure}


Occlusion naturally occurs in both the training and test sets due to the random initialization of object 6D poses. Figure~\ref{fig:occlusion} shows an example from test set, in which our method predicts preparatory actions to expose an initially occluded functional part (the lighter button). Although anticipatory pose prediction becomes less constrained under partial observation, the method still utilizes global geometric structure to infer a plausible preparatory action. After the object is reoriented, Goal Affordance is applied again to localize the functional region more precisely, enabling reliable downstream execution.
\section{Failure Case Analysis} 
\label{supp:fail_analysis}

\begin{figure}[htbp]
\centering
\includegraphics[width=0.9\linewidth]{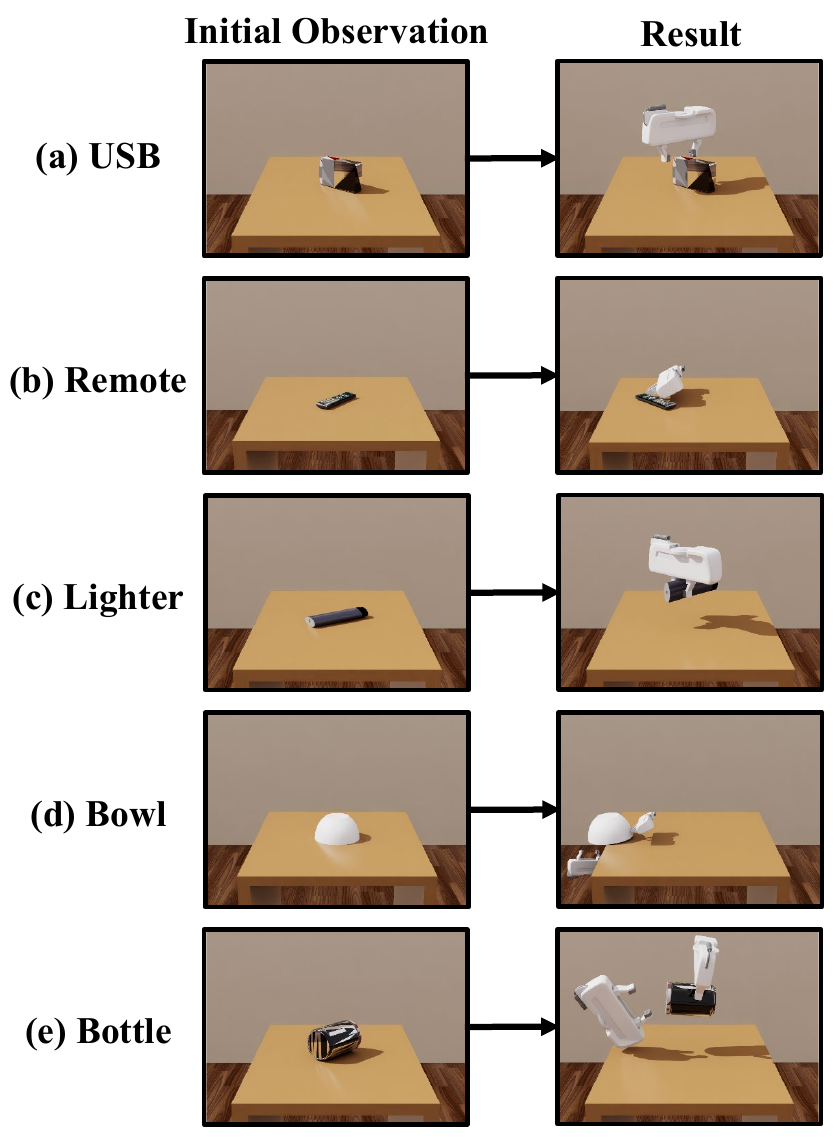}
\caption{
\textbf{Representative failure cases.}
These examples illustrate scenarios in which the robot struggles to determine appropriate actions, reflecting the inherent difficulty of the tasks. 
The first column shows the input observations, and the second column presents the resulting failures.
} 
\label{fig:Failure_analysis}
\end{figure}

Figure~\ref{fig:Failure_analysis} illustrates several common failure modes of our model. Below, we discuss the underlying causes of these errors and outline potential directions for improvement.

(a) Challenges posed by small components.
In this case, the assistant arm attempts to grasp the non-movable body of a USB drive while avoiding its movable cap, as disturbing the cap would impede the subsequent grasp by the primary arm. However, due to the small size of the non-movable body, the assistant gripper tends to approach very close to the bottom edge.
This often results in grasps that are too close to the boundary, yielding an unstable hold.

(b) Constraints of one-shot pushing.
Here, the assistant arm attempts to push the object toward the table edge, but the object becomes unstable during the motion. The primary reason is that our task setting allows only a single pushing attempt. Achieving a stable final pose with a single push is inherently difficult, especially for objects that shift unpredictably. A reasonable mitigation strategy is to allow multiple small corrective pushes, enabling the robot to iteratively adjust its contact points and orientations in a closed-loop manner. However, such iterative pushing strategies fall outside the central focus of this work, which emphasizes inter-arm collaboration. We leave multi-step closed-loop pushing strategies to future work.

(c) Action ambiguity caused by occlusion.
After the primary arm grasps and lifts the lighter, the gripper unintentionally occludes the lighter’s ignition button from the camera’s viewpoint. This occlusion introduces ambiguity when the model predicts the subsequent reorientation action, sometimes resulting in incorrect decisions. This limitation arises because our setup uses only a single camera. A multi-camera setup could substantially reduce such ambiguities. Another potential direction is to teach the policy to reorient objects proactively to reveal occluded regions, similar to strategies used in object-reconstruction tasks~\cite{pfaff2025scalable,dasgupta2024uncertainty,jia2024efficient}.

(d) Geometric constraints near the table surface.
In this case, the primary arm attempts to grasp the rim of a small-diameter bowl but collides with the table surface. We observe a high failure rate for such bowls because their geometry significantly restricts feasible grasp poses, increasing the likelihood of table contact.

(e) Sensitivity of interaction poses.
Objects such as bottles with wide, shallow lids require highly precise grasp poses. Even minor deviations in the gripper’s orientation or position can result in failure. Including several such objects in our benchmark allows for a more comprehensive evaluation of the model’s manipulation precision.


\end{document}